\documentclass[10pt,twocolumn,letterpaper]{article}

\usepackage[pagenumbers]{cvpr} % To force page numbers, e.g. for an arXiv version

\definecolor{cvprblue}{rgb}{0.21,0.49,0.74}
\definecolor{mmfusion}{RGB}{216, 177, 214}
\definecolor{assist}{RGB}{236, 170, 90}
\definecolor{seg}{RGB}{149, 221, 214}
\usepackage[pagebackref,breaklinks,colorlinks,allcolors=cvprblue]{hyperref}

\usepackage{indentfirst}
\usepackage{multirow}
\usepackage{utfsym}
\usepackage{pifont}

\usepackage{amsmath}
\usepackage[linesnumbered,ruled]{algorithm2e}

\usepackage{xspace}
\usepackage{listings}
\newcommand*{\affaddr}[1]{#1} % No op here. Customize it for different styles.
\newcommand*{\affmark}[1][*]{\textsuperscript{#1}}

\newlength\savewidth\newcommand\shline{\noalign{\global\savewidth\arrayrulewidth
\global\arrayrulewidth 1pt}\hline\noalign{\global\arrayrulewidth\savewidth}}

\def\paperID{6171} % *** Enter the Paper ID here
\def\confName{CVPR}
\def\confYear{2025}

\title{Audio-Visual Instance Segmentation}

\author{
Ruohao Guo\affmark[1], Xianghua Ying\affmark[1]\thanks{Corresponding Author} , Yaru Chen\affmark[2], Dantong Niu\affmark[3], Guangyao Li\affmark[4], Liao Qu\affmark[5], Yanyu Qi\affmark[6], \\
Jinxing Zhou\affmark[7] Bowei Xing\affmark[1], Wenzhen Yue\affmark[1], Ji Shi\affmark[1], Qixun Wang\affmark[1], Peiliang Zhang\affmark[8], Buwen Liang\affmark[6] \\[0.6\baselineskip]
\affaddr{\affmark[1]State Key Laboratory of General Artificial Intelligence, School of Intelligence Science and \\ Technology, Peking University}, 
\affaddr{\affmark[2]University of Surrey}, \affaddr{\affmark[3]UC Berkeley}, \affaddr{\affmark[4]Tsinghua University}, \\ \affaddr{\affmark[5]CMU}, 
\affaddr{\affmark[6]China Agricultural University}, \affaddr{\affmark[7]MBZUAI}, \affaddr{\affmark[8]Wuhan University of Technology}
}

\begin{document}
\maketitle
\begin{abstract}
In this paper, we propose a new multi-modal task, termed audio-visual instance segmentation (AVIS), which aims to simultaneously identify, segment and track individual sounding object instances in audible videos. To facilitate this research, we introduce a high-quality benchmark named AVISeg, containing over 90K instance masks from 26 semantic categories in 926 long videos. Additionally, we propose a strong baseline model for this task. Our model first localizes sound source within each frame, and condenses object-specific contexts into concise tokens. Then it builds long-range audio-visual dependencies between these tokens using window-based attention, and tracks sounding objects among the entire video sequences. Extensive experiments reveal that our method performs best on AVISeg, surpassing the existing methods from related tasks. We further conduct the evaluation on several multi-modal large models. Unfortunately, they exhibits subpar performance on instance-level sound source localization and temporal perception. We expect that AVIS will inspire the community towards a more comprehensive multi-modal understanding. Dataset and code is available at \href{https://github.com/ruohaoguo/avis}{https://github.com/ruohaoguo/avis}.
\end{abstract}
\section{Introduction}
\label{sec:intro}

%-------------------------------------------------------------------------
Vision and hearing are our primary channels of communication and sensation \cite{zhou2024towards, zhou2024advancing, zhou2024label, zhou2022contrastive, zhou2021positive, av_event3, guo2024unitr}. Audio-visual collaboration is beneficial for humans to better perceive and interpret the world. Humans have the ability to associate mixed sounds with object instances in complicated realistic scenarios. Imagine a cocktail-party scenario: when a group of people is speaking, we can not only locate the sound sources but also determine how many people are talking.

\begin{figure*}[t]
\centering
\includegraphics[width=1.0\textwidth]{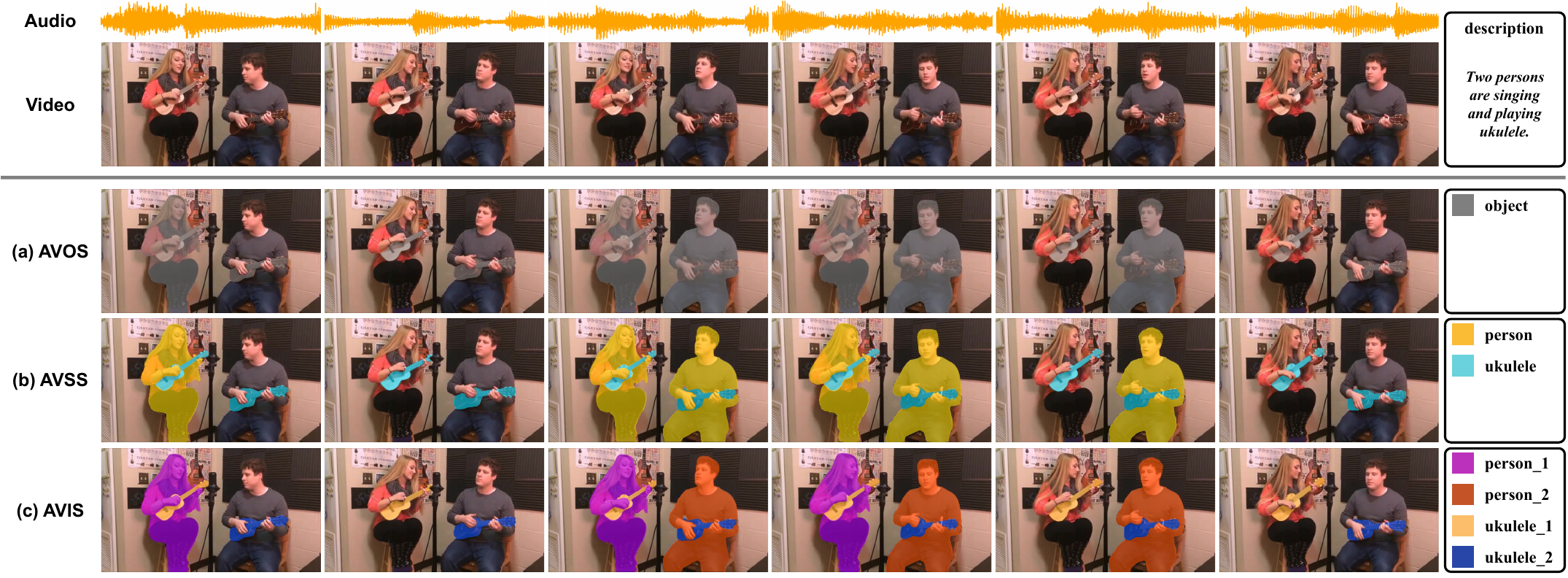}
\vspace{-1.3em}
\caption{Comparison of different audio-visual segmentation tasks. (a) Audio-Visual Object Segmentation (AVOS) only requires binary segmentation. (b) Audio-Visual Semantic Segmentation (AVSS) associates one category with every pixel. (c) Audio-Visual Instance Segmentation (AVIS) treats each sounding object of the same class as an individual instance.}
\label{fig_intro}
\end{figure*}

Inspired by this human perception, we explore instance-level sound source localization in long videos and propose a new task, namely audio-visual instance segmentation (AVIS). As can be seen in Figure \ref{fig_intro} (c), it requests a model to simultaneously classify, segment and track sounding object instances—identify \textit{which object categories are making sounds}, infer \textit{where the sounding objects are}, and monitor \textit{when they are making sounds}. This new task facilitates a wide range of practical applications, including embodied robotics, video surveillance, video editing, etc. Moreover, it can serve as a fundamental task for evaluating the comprehension capabilities of multi-modal large models.

Audio-visual instance segmentation is related to several existing tasks. For example, audio-visual object segmentation (AVOS) \cite{avsbench} is to separate sounding objects from the background region of a given audible video, as shown in Figure \ref{fig_intro} (a). Unlike AVOS being tasked with binary foreground segmentation, audio-visual semantic segmentation (AVSS) \cite{avss} aims at predicting semantic maps that assign each pixel with a specific category, as shown in Figure \ref{fig_intro} (b). To accomplish the above tasks, many works \cite{CATR, avsegformer, combo, guo2024open} extend the image segmentation frameworks \cite{detr, mask2former} to the video domain, and design various audio-visual fusion modules for sound source localization. Despite promising performance in the AVSBench dataset \cite{avss}, current methods still suffer from two limitations in real-world scenarios. First, these methods fail to differentiate two sounding objects with the same category, such as the woman, man, left ukulele and right ukulele depicted in Figure \ref{fig_intro}. Second, these methods focus on 5- or 10-second trimmed short videos and ignore long-range modeling abilities, which may lead to weak performance in real world. 

One potential reason that the AVIS task is rarely studied is the absence of a high-quality dataset. Despite the existence of audio-visual segmentation datasets \cite{avsbench, avss}, none are directly applicable to our proposed task, due to lacking instance-level annotations and long-form videos. Therefore, in this work, we built the first audio-visual instance segmentation dataset, namely AVISeg. The new dataset consists of 926 videos with an average duration of 61.4 seconds and 94,074 high-quality masks, covering 26 common categories from 4 real-world scenarios (Music, Speaking, Machine and Animal). Our dataset can also be served as a benchmark for AVOS and AVSS tasks. Additionally, we present a novel evaluation metric, termed frame-level sound localization accuracy (FSLA), which measures the proportion of frames that are correctly predicted by the model out of the total number of frames.

In order to deal with the above AVIS task, we follow the query-based segmentation paradigm \cite{mask2former, vita} and propose a baseline model called AVISM. To be specific, a frame-level sound source localizer segments sounding objects within each frame independently and summarizes per-frame scenes into a small amount of object tokens. Then, a video-level sounding object tracker is designed to build frame-to-frame communications and track sounding objects throughout the entire video. To lessen computational overheads in processing long and high-resolution videos, the tracker uses the concise object tokens as a mean of conveying information rather than dense image features, and adopts window-based self-attention mechanisms to efficiently capture long-range dependencies in consecutive frames. Experimental results demonstrate the superiority of our baseline. Additionally, we make a thorough evaluation of several prominent multi-modal large models on our AVISeg dataset. Surprisingly, these self-proclaimed large models are far from satisfactory in instance-aware sound source localization and temporal perception. Our dataset emphasizes the necessity for further improvements in handling audio-visual data and long videos, providing insights for future development of multi-modal large models. Our contributions are as follows:

(1) To our best knowledge, this is the first work exploring audio-visual instance segmentation, which aims to classify, segment and track sounding objects in given audible videos.

(2) We create a high-quality video dataset to support the above task, containing 926 videos with an average length of 61.4s. Besides, we propose a novel frame-level metric for evaluating audio-visual instance segmentation.

(3) A strong baseline model is developed to localize sound source in each frame and track sounding objects in the entire video. To handle long videos, it distils image features into a small number of tokens and uses window-based attention to convey audio-visual temporal information.

(4) Extensive experiments indicate that our framework achieves state-of-the-art results under all evaluation metrics. Moreover, our dataset can also serve as a potential benchmark for evaluating various multi-modal large models.

\section{Related Work}
\label{sec:rw}

%-------------------------------------------------------------------------
\subsection{Video Instance Segmentation}
Video instance segmentation (VIS) aims at simultaneous segmentation and tracking of all object instances in videos. Early methods \cite{vis, Stem_seg, SipMask, sgnet, CrossVIS} often extend CNN-based image segmentation methods \cite{Mask_RCNN, sotr} to establish temporal consistency. For example, MaskTrack R-CNN \cite{vis} introduces an additional tracking head to Mask R-CNN \cite{Mask_RCNN} for object matching and association between frames. SG-Net \cite{sgnet} follows the anchor-free FCOS detector \cite{FCOS} and directly leverages the object centerness from detection to delineate the temporal coherence in video sequences. The above approaches require extra post-processing steps, such as non-maximal suppression (NMS), leading to higher computational costs and potential misdetections. Recent methods \cite{vistr, mask2former2, tevit, seqformer, ifc, vita, instanceformer, dvis} adapt Transformer-based image segmentation methods \cite{detr, mask2former} to the VIS task. For example, VisTR \cite{vistr} builds on the query-based DETR \cite{detr} and naturally outputs the sequence of masks for each instance without heuristic matching or hand-designed post-processing. Follow-up works, such as Mask2Former-VIS \cite{mask2former2} and SeqFormer \cite{seqformer}, design more querying strategies to improve the performance of segmentation and tracking. To avoid heavy computation and memory usage, IFC \cite{ifc} and VITA \cite{vita} first distill dense spatio-temporal features into a small amount of tokens, and then perform inter-frame communication between tokens. This information-passing paradigm allows models for efficiently handling long and high-resolution videos with a common GPU.

\subsection{Audio-Visual Segmentation}
Audio-visual segmentation (AVS) focuses on localizing and segmenting sounding objects within each video frame. Zhou et al. \cite{avsbench, avss} introduce the first AVS dataset, namely AVSBench, which serves two different sub-tasks including audio-visual object segmentation (AVOS) and audio-visual semantic segmentation (AVSS). The former \cite{avsbench} requires producing binary masks of sounding objects, while the latter \cite{avss} further needs to generate semantic maps representing the object category. To address these problems, they employ a standard encoder-decoder architecture with a modified non-local block to encode space-time relation and segment sounding objects. CAVP \cite{CAVP} builds an AVS dataset by randomly matching the images from COCO \cite{coco} and audio files from VGGSound \cite{Vggsound} based on the semantic classes of the objects. Inspired by DETR \cite{detr} and Mask2Former \cite{mask2former}, recent works \cite{avsegformer, combo, CATR, guo2024open, guo2024instance} adopt the query-based architecture decode masks for sounding objects. For example, AVSegFormer \cite{avsegformer} trivially incorporates audio features and learnable queries, enabling the decoder to capture relevant visual semantics and predict the audio-constrained masks. COMBO \cite{combo} explores multi-order bilateral relations in modality, temporal and pixel levels for the AVSS task. Notably, a bilateral-fusion module is designed to align audio and visual modalities bi-directionally and assist the model in segmenting the sounding objects.

\begin{figure*}[htp]
\centering
\includegraphics[width=1.0\textwidth]{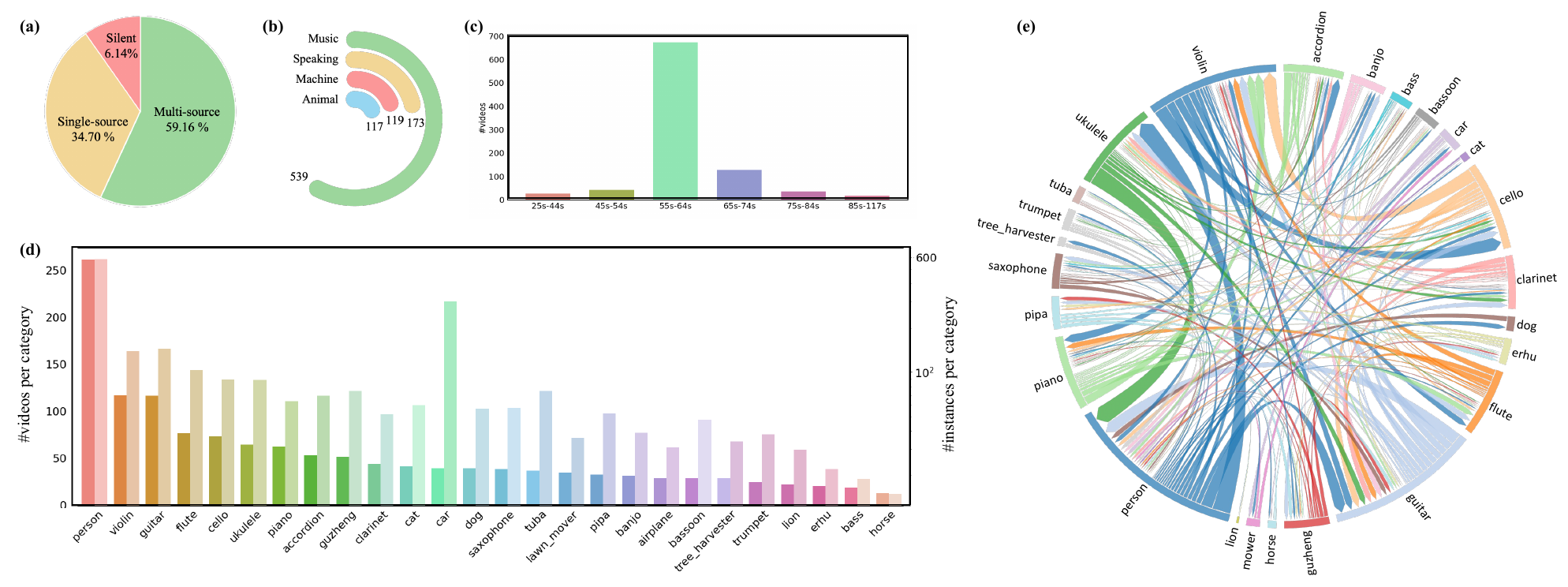}
\vspace{-1.3em}
\caption{Illustrations of our AVISeg dataset statistics. (a) Ratio of different sound sources. (b) Number of video in 4 real-world scenarios. (c) Distribution of video lengths. (d) Number of video and objects for the 26 categories. (e) Relations between different categories.}
\label{fig_statistics}
\end{figure*}

\section{New Task}
\label{sec:task}

%-------------------------------------------------------------------------
\subsection{Problem Definition}
Audio-visual instance segmentation (AVIS) is a challenging multi-modal task that involves localizing and segmenting sounding objects in a video, while assigning each a unique identity label to ensure consistent tracking throughout the video. In this task, we predefine a category label set as $\mathcal{C} = \{1, ..., K\}$, where $K$ is the number of categories. Given a video sequence with $T$ frames and its corresponding audio, suppose there are $N$ sounding objects belonging to the category label set $\mathcal{C}$ in the video. For each sounding object $o^i$, let $c^i \in \mathcal{C}$ denote its category label, and let $m^i_t$ denote its binary segmentation mask in $t^{th}$ frame where $t \in T^{\prime}$ and $T^{\prime}$ denotes the sounding time set, i.e., $T^{\prime} \subseteq T$. We assume that an AVIS model outputs $H$ instance hypotheses. For each hypothesis $o^j$, it needs to contain a predicted category label $\tilde{c}^j \in \mathcal{C}$, a confidence score $\tilde{s}^j \in [0,1]$, and a sequence of predicted binary masks $\tilde{m}^j_{\tilde{t}}$. The goal of AVIS task is to minimize the difference between the ground truth and the hypotheses. This requires the AVIS model to correctly determine which instances are making sounds, accurately identify and segment these sounding instances, and reliably track them in the entire video.

\subsection{Evaluation Metrics}
To evaluate how well an AVIS model performs, we need to choose appropriate metrics to compare its outputs with the ground truth. In our task, we adopt two evaluation protocols including the mean Average Precision (mAP) \cite{vis} and the Higher Order Tracking Accuracy (HOTA) \cite{hota}. mAP follows the computation of the average precision-recall metric over trajectories, which is commonly used in video instance segmentation. However, mAP is not perfectly suited to our task, because it can be increased by producing many different predictions with low confidence scores and does not decrease even if non-sounding objects are predicted. HOTA performs a bijective matching at the detection level while scoring association over trajectories, which is designed for multi-object tracking task. This makes HOTA a balanced metric for measuring both detection and association. When applied to the AVIS task, it can penalize those models that predict non-sounding objects.

Besides considering the above object-based metrics, we propose a novel measure, namely frame-level sound localization accuracy (FSLA), tailored to measure the proportion of frames that are correctly predicted by the model out of the total number of frames. Specifically, we first use the Hungarian algorithm \cite{kuhn1955hungarian} to determine a one-to-one matching between ground-truth and predicted detections. For each frame, it can be treated as correct frame if it satisfies the following conditions: 1) The number of sounding objects is correct; 2) The category of the sounding objects is correct; 3) The IoU (Intersection over Union) between the ground truth and the predicted sounding objects is greater than threshold $\alpha$. The final score is computed by averaging over all classes before averaging different $\alpha$ thresholds (0.05 to 0.95 in 0.05 intervals). The pseudo code of the FSLA metric is in the \textit{Supp. Materials}. Compared to other metrics, our FSLA allows for easier localization of incorrect frames and offers a more intuitive explanation of the model's performance across different time periods. Additionally, it can be decomposed into a set of sub-metrics (FSLAn, FSLAs and FSLAm) which can be used for model evaluation in scenarios with no sound source, a single sound source, and multiple sound sources. This results in FSLA being able to guide how models can be improved, or understand where they are likely to fail when used.

\section{Dataset}
\label{sec:dataset}

% 1) Video Collection
To explore audio-visual instance segmentation and evaluate the proposed methods, we create a new large-scale benchmark called AVISeg. Considering that this task involves complex audio-visual interactions and requires high-quality data, we manually collect and choose 926 videos from YouTube and the publicly available datasets \cite{music-avq, li2024object, li2024patch}, e.g., MUSIC-AVQA. Our released AVISeg dataset satisfies the following criteria: 1) It focuses on long-term videos (61.4s), bringing them much closer to real applications. 2) It contains 26 common sound categories, spanning 4 dynamic scenarios: ``Music'', ``Speaking'', ``Machine'', and ``Animal''. 3) It involves some challenging cases, such as videos with silent sound sources, single sound source, and multiple sources simultaneously. These attributes impose higher demands on the model for accurate recognition, segmentation, and tracking of sounding objects.

% 2) Annotation
Similar to AVSBench \cite{avsbench}, each video is divided into 1-second clips. We then adopt an interactive semi-automatic annotation tool \footnote[1]{\footnotesize\url{https://www.yatenglg.cn/isat/}} based on ViT-H SAM model \cite{sam} to label sounding object instances belonging to the defined category set exhaustively in these videos. For example, in the first column of Figure \ref{fig_intro}, the woman is labeled as ``person\_1'' because she is singing, while the man is not labeled since no sound is made. That is, an object will only be masked and assigned a unique identifier when it emits sound. Note that each labeled frame undergoes multiple rounds of manual review and refinement to ensure high-quality annotations.

% 3) Dataset Statistics
In terms of high-level statistics, our AVISeg dataset consists of 94,074 masks on 56,871 frames, distributed in 926 videos for about 16 hours. Figure \ref{fig_statistics} (a-e) provides the statistical analysis of our dataset. In this dataset, silent frames, single-source frames and multi-source frames account for 6.14\%, 34.70\% and 59.16\%, respectively. AVISeg covers 4 real-world scenarios, with the ``Music'' scenario having the largest number of videos, totaling 539. Note that a video may belong to multiple scenarios, such as the simultaneous appearance of animals and musical instruments. A comparison of the proposed AVISeg and related datasets is shown in Table \ref{tab: dataset}. For training and evaluation, we randomly split the dataset into training, validation, and testing sets with 616, 105, and 205 videos, respectively.

\begin{table}[htp]
\caption{Comparison with other datasets from related tasks. SSL represents audio-visual event localization.}
\label{tab: dataset}
\resizebox{\linewidth}{!}{
\begin{tabular}{c|cccccc}
\shline
Task & Dataset & Videos  & Length  & Classes & Anno. \\

\hline
\multirow{2}{*}{SSL} 
& Flickr-S \cite{flickr-s}  & 5,000   & 20.0s     & 50 & bbox  \\
& VGG-SS \cite{vgg-ss}  & 5,158   & 10.0s    & 220      & bbox  \\

\hline
\multirow{1}{*}{AVOS} 
& AVSBench-O \cite{avsbench}  & 5,356    & 5.0s     & 23 & pixel  \\

\hline
\multirow{1}{*}{AVSS} 
& AVSBench-S \cite{avss}  & 12,356   & 7.8s  & 70 & pixel  \\

\hline
\multirow{2}{*}{VIS} 
& YTVIS \cite{vis}  & 2,883  & 4.6s  & 40 & pixel  \\
& OVIS \cite{ovis}  & 901    & 12.8s & 25 & pixel  \\

\hline
\multirow{1}{*}{AVIS} 
& AVISeg    & 926    & 61.4s    & 26    & pixel \\
\shline
\end{tabular}}
\end{table}

\section{Baseline Model}
\label{sec:method}

%-------------------------------------------------------------------------

\begin{figure*}[!htp]
\centering
\includegraphics[width=1.0\textwidth]{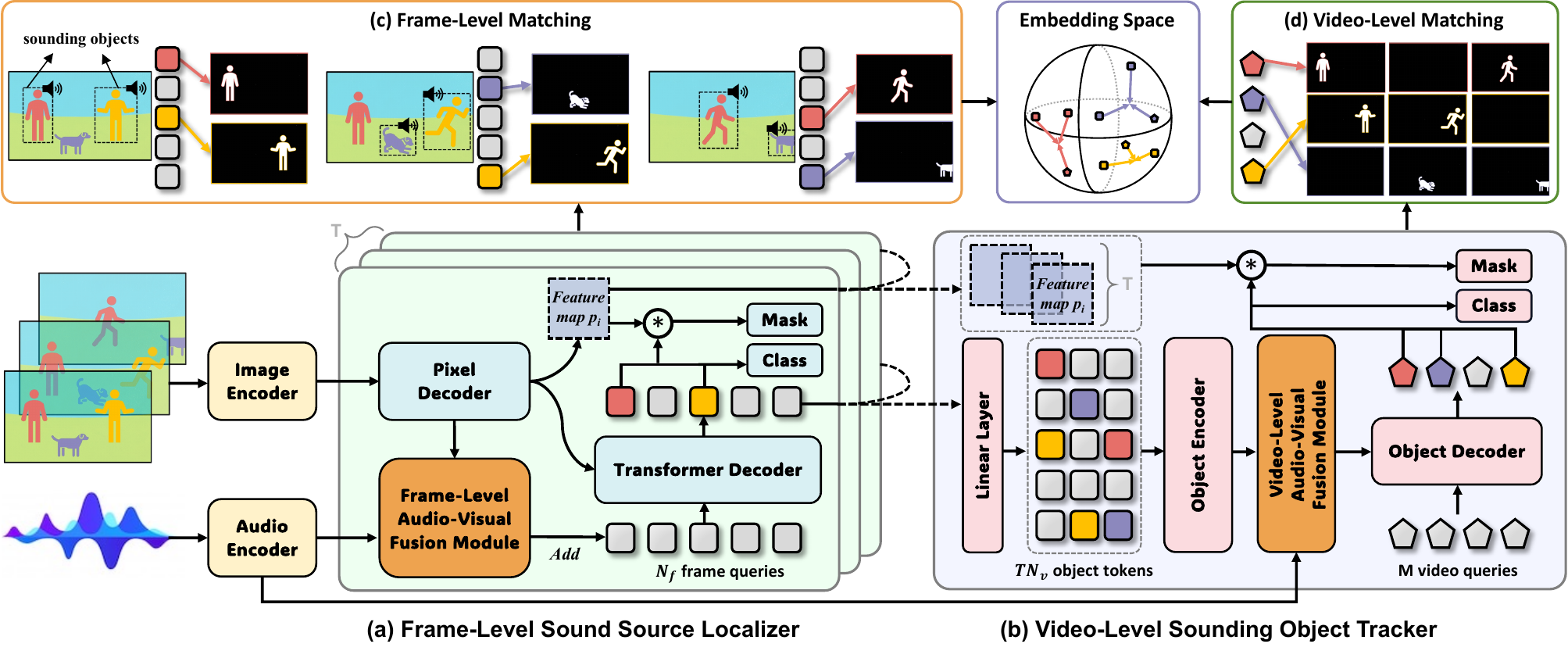}
\put(-242,6.5){\footnotesize ~(\S\ref{sec:fl})}
\put(-40,6.5){\footnotesize ~(\S\ref{sec:vl})}
\caption{Overview of the proposed AVISM for audio-visual instance segmentation. (a) The frame-level sound source localizer segments sounding objects within each frame independently and condenses dense image features into frame queries. (b) The video-level sounding object tracker takes frame queries and audio features as input, and then performs temporal audio-visual communications between frames.}
\label{fig_model}
\end{figure*}

We introduce a new baseline model, termed AVISM, for the audio-visual instance segmentation task. The proposed AVISM model, built upon Mask2Former \cite{mask2former, mask2former2} and VITA \cite{vita}, adopts a query-based Transformer architecture to learn a set of query vectors representing sounding objects for the instance segmentation and tracking. To better model audio-visual semantic correlations in long and complicated videos, we present the frame-level audio-visual fusion module and video-level audio-visual fusion module to integrate audio and visual features. The overall framework of our baseline model is illustrated in Figure \ref{fig_model}.

\subsection{Audio-Visual Representation}
Given an input video sequence that contains both visual and audio tracks, we split it into $T$ non-overlapping visual and audio snippet pairs $\{V, A\}=\{v_i, a_i\}_{i=1}^T$, where each snippet spans 1 second and $T$ represents the number of snippets as well as the video length. For each visual snippet $v_i$, we apply ResNet \cite{resnet} or Swin Transformer \cite{swin} as the backbone to extract hierarchical features $f_{i,k}^V \in \mathbb{R}^{H_k \times W_k \times D_k}$. $H_k \times W_k$ denotes the output resolution of each $v_i$ at the $\text{k}^{\text{th}}$ backbone level. The final visual representation can be formulated as $F^V=\{f_i^V\}_{i=1}^T$. For each audio snippet $a_i$, we first convert it to a mel spectrogram via the short-time Fourier transform and then encode it into an audio feature vector $f_i^A \in \mathbb{R}^D$ using a pre-trained VGGish model \cite{vggish}, where $D$ is the feature dimension. The final audio representation $F^A=\{f_i^A\}_{i=1}^T$ is extracted offline and the VGGish model is not fine-tuned during the training process.

\subsection{Frame-Level Sound Source Localizer} \label{sec:fl}
To accurately localize the sounding objects within each video frame, we propose the frame-level sound source localizer that establishes the spatial association between audio and visual modalities. As depicted in Figure \ref{fig_model} (a), we employ a multi-scale deformable attention Transformer \cite{zhu2020deformable}, namely pixel decoder, to produce enhanced visual features $\hat{f}_i^V$ and high-resolution per-pixel embeddings $p_i$. Then, the frame-level audio-visual fusion module performs cross-attention computation between $\hat{f}_i^V$ and the corresponding audio feature $f_i^A$ at multiple scales, yielding audio-to-image features $f_i^{AV} \in \mathbb{R}^{C}$. Inspired by the set prediction paradigm \cite{detr}, we introduce $N_f$ audio-conditioned learnable queries, which are added with $f_i^{AV}$ to form \textit{frame queries} $Q_f \in \mathbb{R}^{N_f \times C}$. After a Transformer decoder distills and embeds visual semantics of all frames into the frame queries, each frame query is dot-multiplied with $p_i$, and used for classifying and segmenting its matched sounding object.

\subsection{Video-Level Sounding Object Tracker} \label{sec:vl}
One limitation of the above localizer is that it operates independently on each frame, with no inter-computation shared across frames. For the solution to this problem, we present the video-level sounding object tracker that builds temporal communications throughout the entire video sequence. Considering the heavy computation demands posed by processing long and high-resolution videos, our tracker takes the frame queries as inputs rather than image features, and leverages the window-based self-attention mechanisms \cite{swin} to capture long-range dependencies among frames. 

\begin{figure}[!htp]
\centering
\includegraphics[width=1.0\linewidth]{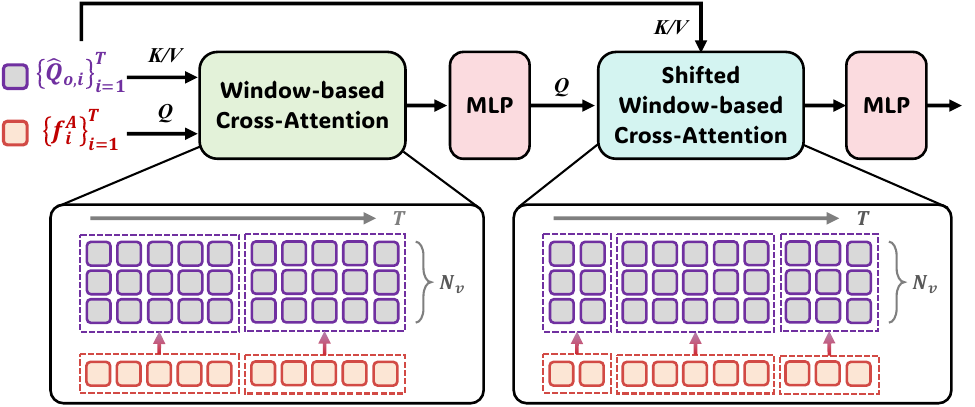}
\caption{The architecture of our proposed video-level audio-visual fusion module. For the entire video sequence, it computes cross-attention between object tokens $\{\hat{Q}_{o,i}\}_{i=1}^T$ and audio features $\{f_i^A\}_{i=1}^T$ within local windows, and introduces cross-window connections by shifting windows.}
\label{fig_avfusion}
\end{figure}

As shown in Figure \ref{fig_model} (b), a linear layer converts $T \times N_f$ frame queries gathered from all frames into object tokens $Q_o$. The object encoder, similar to \cite{vita}, partitions these object tokens along the temporal axis into non-overlapping local windows of size $W$, within which self-attention is performed. After alternately shifting the windows, object tokens $\hat{Q}_o$ from different windows can exchange object-wise information. We extend this capability of processing long videos to multi-model temporal learning, and design a video-level audio-visual fusion module (Figure \ref{fig_avfusion}) incorporating $N$ attention layers. In each local window, it calculates cross-attention between object tokens $\hat{Q}_o$ and audio features $f_i^A$. As the local window shifts and the attention layer goes deeper, our model can efficiently achieve frame-to-frame audio-visual communications in long videos. Its outputs are added with $\hat{Q}_o$ and their results are referred as $Q_o^{AV}$. This temporal fusion benefits the global alignment of audio and object instances, while also enhancing object tracking and identity association across different frames.

To decode object-centric information from all object tokens, we initialize a fixed set of learnable \textit{video queries} $Q_v \in \mathbb{R}^{N_v \times C}$, where $N_v$ is the number of video queries. The object decoder, implemented as a standard Transformer decoder \cite{detr, vita}, receives $Q_o^{AV}$ and aggregates their semantics into video queries. At the end of the decoder, two output heads are exploited to obtain the final predictions, with each head comprising two fully-connected layers. Specifically, a class head predicts class probabilities $p \in \mathbb{R}^{K+1}$ for each video query, including a \textit{no sounding object} $\varnothing$ class in addition to the $K$ given classes of a dataset. Besides, object queries are input into a mask head and then dot-multiplied with $p_i$, resulting in the final mask logits.

\subsection{Training Loss} \label{sec:tl}
There are three terms in the training loss as follows:
\begin{equation}
\mathcal{L} = \lambda_{\text{frame}} \mathcal{L}_{\text{frame}} + \lambda_{\text{video}} \mathcal{L}_{\text{video}} + \lambda_{\text{sim}} \mathcal{L}_{\text{sim}}
\end{equation}
where $\lambda_{\text{frame}}$, $\lambda_{\text{video}}$ and $\lambda_{\text{sim}}$ are hyper-parameters to balance the loss terms. Their default values are set to 1, 1, 0.5, respectively. For frame-wise supervision, we first compute costs between frame queries and ground truth at each $t^{\text{th}}$ frame using the cost function of Mask2Former \cite{mask2former}. Following DETR \cite{detr}, the Hungarian algorithm \cite{kuhn1955hungarian} is then employed for optimal matching, as shown in Figure \ref{fig_model} (c). Finally, we utilize $\mathcal{L}_{\text{frame}}$ from \cite{mask2former} to calculate loss between the matched pairs. For video-wise supervision, we also search for optimal assignment between video queries and ground-truth sequences using the cost function of IFC \cite{ifc}, as shown in Figure \ref{fig_model} (d). These bipartitely matched pairs are used to compute the loss function $\mathcal{L}_{\text{video}}$ from \cite{ifc}, a simple extension of \cite{mask2former}. Additionally, as depicted in Figure \ref{fig_model} (e), we introduce the similarity loss \cite{vis, vita} to align frame queries with video queries in the embedding space, annotating pairs of equal identities as 1 and others as 0.

\section{Experiment}
\label{sec:experiment}

%-------------------------------------------------------------------------
\begin{table*}[htp]
\caption{Quantitative evaluation of different models from related tasks on the AVISeg test set. The best results are highlighted in bold.}
\label{tab: mr}
\resizebox{\textwidth}{!}{
\begin{tabular}{c|ccc|ccc|ccccc}
\shline
Task & Model   & Venue  & Audio & FSLA & HOTA & mAP  & FSLAn & FSLAs  & FSLAm &  AssA & DetA \\

\hline
\multirow{5}{*}{VIS} 
& Mask2Former-VIS \cite{mask2former2}  & CVPR' 22 & \usym{2718} & 29.75 & 52.03 & 28.66 & 0.00 & 25.47 & 36.37  & 64.49 & 43.33 \\

& TeViT \cite{tevit}   & CVPR' 22 & \usym{2718} & 32.28 & 53.67 & 31.52 & 0.00 & 28.07 & 39.18  & 65.27 & 45.10 \\

& SeqFormer \cite{seqformer}   & ECCV' 22 & \usym{2718} & 30.32 & 54.32 & 32.79 & 25.03 & 21.76 & 36.46  & 67.25 & 45.23 \\

& VITA \cite{vita}   & NeurIPS' 22 & \usym{2718} & 38.04 & 57.48 & 36.25 & 15.04 & 27.98 & 47.45  & 69.86 & 48.96 \\

& DAVIS \cite{dvis}   & ICCV' 23 & \usym{2718} & 23.99 & 49.12 & 19.83 & 14.61 & 24.83 & 24.69  & 63.51 & 40.11 \\

& LBVQ  \cite{lbvq}  & TCSVT' 24 & \usym{2718} & 34.73 & 56.97 & 36.58 & 27.71 & 29.52 & 38.96  & 68.34 & 48.83 \\

\hline
\multirow{2}{*}{AVSS} 
& AVSegFormer \cite{avsegformer} & AAAI' 24 & \usym{2714} & 35.66 & 55.74 & 35.72 & 18.58 & 27.51 & 43.08  & 67.13 & 48.51  \\
& COMBO  \cite{combo}     & CVPR' 24 & \usym{2714} & 39.49 & 57.39 & 37.84 & 21.91 & 27.18 & 49.63  & 68.87 & 50.12 \\

\hline
\multirow{1}{*}{AVIS} 
& AVISM & CVPR' 25 & \usym{2714} & \textbf{42.78} & \textbf{61.73} & \textbf{40.57} & \textbf{32.22} & \textbf{29.83} & \textbf{52.40} & \textbf{71.15} & \textbf{54.97} \\
\shline
\end{tabular}}
\end{table*}

\subsection{Main Results}
We compare AVISM with the state-of-the-art methods from two related tasks, including video instance segmentation (VIS) and audio-visual semantic segmentation (AVSS). For the VIS methods \cite{mask2former2, tevit, seqformer, vita, instanceformer, dvis, lbvq}, only video frames are used for training, while the audio is disregarded. For the AVSS methods \cite{avsegformer, combo}, they follow the query-based detection paradigm \cite{detr} and achieve instance-level segmentation without altering the model, losses and training procedure. To make the evaluation fair, all methods utilize ResNet-50 pre-trained on ImageNet \cite{imagenet} as the backbone and are trained on the AVISeg dataset for 48,000 iterations.

Table \ref{tab: mr} presents the comparison results, including three main metrics (FSLA, HOTA, mAP) and five sub-metrics (FSLAn, FSLAs, FSLAm from FSLA; AssA, DetA from HOTA). It is worth noting that our AVISM achieves the best results under all evaluation metrics. Compared to the VIS methods, AVISM incorporates audio information and leverages multi-modal contexts to localize sounding objects within video frames, which outperforms the best VITA \cite{vita}. This multi-sensory perception helps to guide our model to determine whether or which objects are making sounds. Compared to the AVSS methods, AVISM condenses per-frame scenes into a small number of frame queries and then establishes inter-frame audio-visual communication between them. Our experimental results demonstrate that using the concise frame queries, instead of dense spatio-temporal features, not only improves AVIS performance but also provides robust practicality for processing long and high-resolution videos. Furthermore, the results confirm the viability of AVISeg as a benchmark for AVIS task.

\begin{figure*}[htp]
\centering
\includegraphics[width=1.0\textwidth]{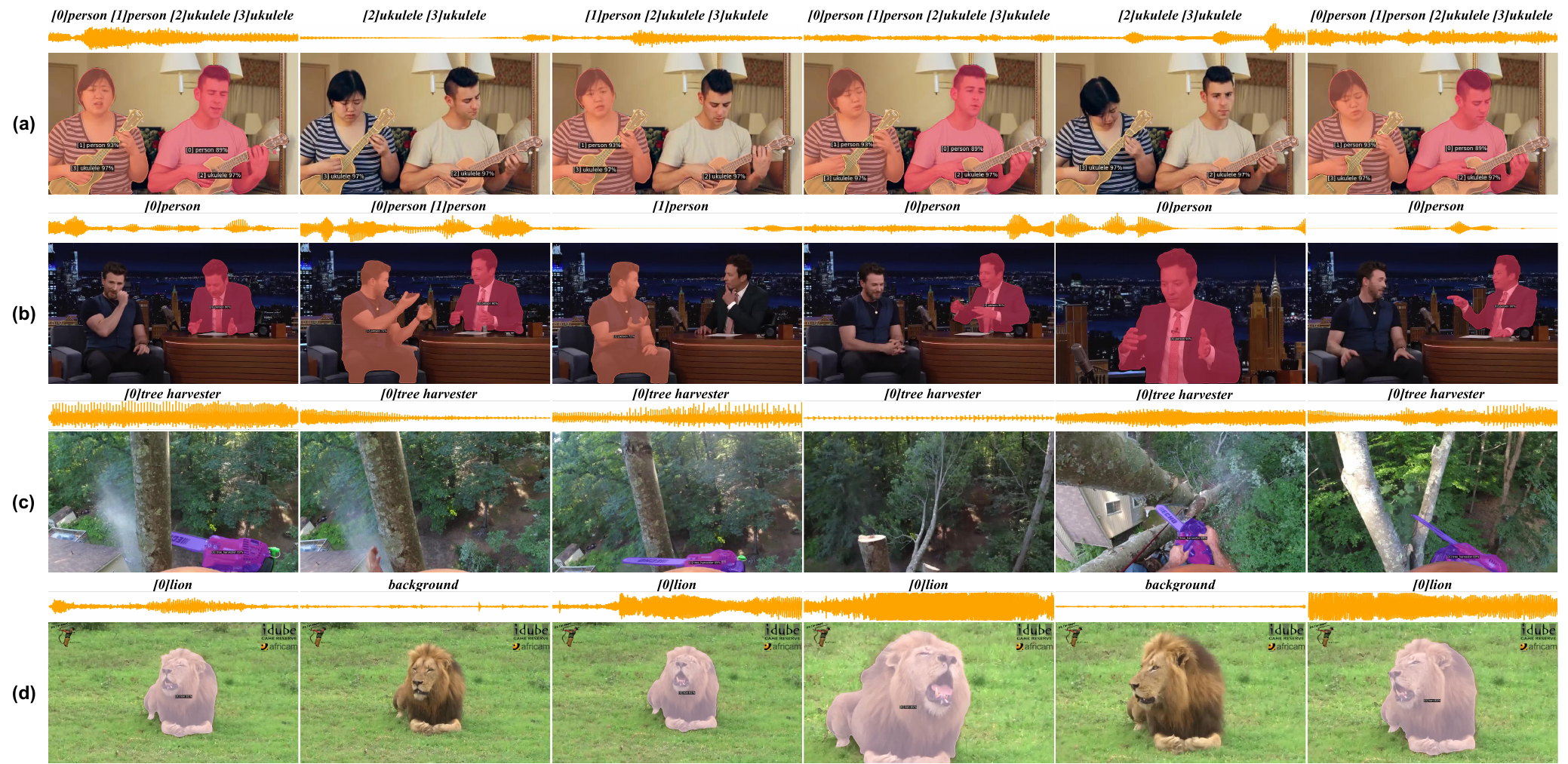}
\caption{Sample results of our baseline model on AVISeg dataset from four scenarios: (a) Music; (b) Speaking; (c) Machine; (d) Animal. Each row have six sampled frames from a video sequence. Zoom in to see details.}
\label{fig_results}
\end{figure*}

Figure \ref{fig_results} visualizes some sample videos with our predictions. Our AVISM model accurately localize the sounding object across both spatial and temporal dimensions, e.g., ``lion'' in video (d). In complex scenes with multiple sound sources, our model enables to handle the numerous mixed semantics, e.g., ``person'' and ``ukulele'' in video (a). When an object begins producing sound in the intermediate frames, AVISM is able to segment it and assign a new identity, as evidenced in video (b). This case also shows the effectiveness of our model in identifying and distinguishing objects with similar appearances or sounds. Moreover, if a sounding object disappears and reoccurs, the AVISM still correctly tracks it, e.g., ``tree harvester'' in video (c).

\begin{table}[htp]
\begin{center}
\caption{Zero-shot results of different multi-modal large models for audio-referred visual grounding on the AVISeg test set.}
\label{tab: mml}
\resizebox{\linewidth}{!}{
\begin{tabular}{cc|ccc}
\shline
Model                                & Assistant & FSLA    & HOTA  & mAP   \\
\hline
Sam4AVS \cite{sam4avs}               & -         & 0.00 & 8.18 & 3.93 \\
BuboGPT-7B \cite{bubogpt}               & GPT-4     & 7.75 & 20.16 & 5.76 \\
PG-Video-LLaVA-7B \cite{pg-video-llava}  & GPT-3.5   & 9.15 & 22.86  & 5.94 \\
AL-Ref-SAM 2 \cite{al-ref-sam}       & GPT-4     & \textbf{18.55}  & \textbf{38.02} & \textbf{15.84} \\
\shline
\end{tabular}}
\end{center}
\end{table}

\subsection{Evaluations on Multi-modal Large Models}

Table \ref{tab: mml} presents the zero-shot results between different multi-modal large models (MMLMs) on AVIS task, revealing that these methods are underperforming. For instance, BuboGPT \cite{bubogpt} and PG-Video-LLaVA \cite{pg-video-llava} localize sound sources with audio-image-text aligned large language models (Vicuna \cite{vicuna} and LLaVA \cite{llava}), and then classifies and segments sounding objects using an off-the-shelf grounding pipeline based on GPT \cite{gpt4} and SAM \cite{sam}. However, BuboGPT is limited to processing a single image and one-second audio, and PG-Video-LLaVA cannot determine the exact time intervals for each sounding object. AL-Ref-SAM 2 \cite{al-ref-sam} adopts Chain-of-Thought prompts to unleash GPT’s temporal-spatial perception and reasoning capabilities. Although pre-trained on large-scale datasets and yielding promising results on audio-visual understanding task, these MMLMs fall short in instance segmentation and long-range modeling, resulting in poor performance on AVISeg. Our new task can provide deeper insights for multi-modal instruct tuning of MMLMs, has the potential to serve as a benchmark for evaluating their performance. More analysis can be found in \textit{Supp. Materials}.

\subsection{Ablation Studies}
\textbf{Impact of audio-visual fusion modules.}
To evaluate our proposed frame-level audio-visual fusion module (FL-AVFM) and video-level audio-visual fusion module (VL-AVFM), we first establish a baseline by disabling both modules. As evidenced in Table \ref{tab: ab1}, the introduction of FL-AVFM yields substantial improvements across all metrics. These gains underscore the importance of effective audio-visual information aggregation at the frame level for enhancing per-frame object localization accuracy.
Further incorporation of the VL-AVFM leads to more pronounced enhancements across all metrics, with the full configuration achieving optimal results. This observation suggests that the VL-AVFM plays a crucial role in leveraging temporal information across frames, thereby facilitating improved tracking consistency and accuracy. Our findings support the hypothesis that temporal audio-visual fusion is instrumental in resolving ambiguities during object tracking, particularly in challenging scenarios where motion cues may be insufficient for determining whether an object is producing sound. This demonstrates the potential of audio as auxiliary information to guide audio-visual instance segmentation.

\begin{table}[htp]
\small
\begin{center}
\caption{Impact of frame-level audio-visual fusion module (FL-AVFM) and video-level audio-visual fusion module (VL-AVFM).}
\label{tab: ab1}
\begin{tabular}{cc|ccc}
\shline
FL-AVFM     & VL-AVFM     & FSLA    & HOTA  & mAP   \\
\hline
            &             & 38.04 & 57.48 & 36.25 \\
\usym{2714} &             & 39.68 & 59.59 & 39.06 \\
\usym{2714} & \usym{2714} & \textbf{42.78} & \textbf{61.73}  & \textbf{40.57} \\
\shline
\end{tabular}
\end{center}
\end{table}
\vspace{-1em}

\textbf{Impact of local window size within video-level sounding object tracker.}
Table \ref{tab: ab2} presents an ablation study on local window sizes in our video-level sounding object tracker. We observe a clear trade-off between the maximum number of processable frames and tracking performance. A window size of 3 allows processing of the longest sequences (5304 frames) but yields the lowest performance across all metrics. Conversely, a window size of 12 significantly improves tracking accuracy at the cost of reduced frame capacity (1416 frames). The performance gain can be attributed to the expanded temporal receptive field, which allows the model to capture more complex inter-frame dependencies. This enhanced temporal context enables the tracker to better understand the long-term dynamics of sounding objects, leading to more accurate localization and tracking. Considering the trade-off between segmentation performance and the ability to process longer sequences, we chose a window size of 6 as the default, which provides a balanced compromise between accuracy and frame capacity.

\begin{table}[htp]
\small
\begin{center}
\caption{Impact of local windows size within video-level sounding object tracker. The maximum number of frames is reported on a single NVIDIA Quadro 6000 GPU.}
\label{tab: ab2}
\begin{tabular}{c|c|ccc}
\shline
Window Size & Max Frames & FSLA    & HOTA  & mAP   \\
\hline
3           & \textbf{5304}       & 40.83 & 61.13 & 40.14 \\
6           & 2778       & 42.78 & 61.73 & 40.57 \\
12          & 1416       & \textbf{42.96} & \textbf{62.82} & \textbf{41.31} \\
\shline
\end{tabular}
\end{center}
\end{table}
\vspace{-1em}

\textbf{Impact of visual backbone and pre-training dataset.}
We further investigate whether providing a stronger backbone and more pre-training data can further improve the model's AVIS performance. As shown in Table \ref{tab: ab3}, adopting the strategy from Mask2Former \cite{mask2former} that using COCO for additional pre-training of our visual backbone resulted in improvements across all metrics. However, when further fine-tuned on the video instance segmentation dataset OVIS \cite{ovis}, despite an increase in mAP, we observe a slight decrease in FSLA. This is likely because OVIS primarily targets improving the model's video segmentation capabilities, leading to the segmentation of many non-sounding objects, thus not achieving better FSLA scores. Consequently, we opt for the IN+COCO pre-trained visual backbone for subsequent experiments. Replacing the backbone with Swin-L achieves the highest scores across all metrics.

\begin{table}[htp]
\begin{center}
\caption{Impact of visual backbone and pre-training dataset.}
\label{tab: ab3}
\resizebox{\linewidth}{!}{
\begin{tabular}{c|c|c|ccc}
\shline
Backbone    & Pre-trained Datasets  & Param.     & FSLA    & HOTA  & mAP   \\
\hline
\multirow{3}{*}{R-50}   & IN                    & \multirow{3}{*}{527.3}       & 42.78 & 61.73 & 40.57 \\
   & IN+COCO               &       & 44.42 & 64.52 & 45.04 \\
   & IN+COCO+OVIS          &       & 43.68 & 64.64 & 45.76 \\
\hline
\multirow{1}{*}{R-101}  & IN+COCO               & \multirow{1}{*}{599.5}       & 45.06 & 64.80 & 46.61 \\
\hline
\multirow{1}{*}{Swin-L}      & IN+COCO                    & \multirow{1}{*}{1181.8}      & \textbf{52.49} & \textbf{71.13} & \textbf{53.46}  \\
\shline
\end{tabular}}
\end{center}
\end{table}
\vspace{-1em}

\section{Conclusion}

This paper introduces a new task of audio-visual instance segmentation with the goal of identifying, segmenting and tracking individual sounding object instances in videos. We present a high-quality dataset and a strong baseline model, providing some early explorations towards this task. In addition, we evaluate the zero-shot performance of several multi-modal large models, but they are far from satisfactory in instance-level sound source localization and long-range temporal perception. These findings underscore the need for further advancements in fine-grained and time-sensitive instruction tuning. We believe our task will innovate the community on new research ideas and directions for multi-modal understanding, and our dataset has the potential to serve as a platform for testing large models. 

\noindent \textbf{Acknowledgments.} This work was supported by the National Natural Science Foundation of China (NSFC) under Grant No. 62371009, and Beijing Natural Science Foundation under Grant No. L247029.

\newpage
{
    \small
    \bibliographystyle{ieeenat_fullname}
    \bibliography{main}
}

\clearpage
\setcounter{page}{1}
\maketitlesupplementary
\appendix

\section{Overview}

\textbf{\ref{sec:em}} Evaluation Metrics

\textbf{\ref{sec:sup_id}} Implementation Details

\textbf{\ref{sec:mas}} More Ablation Studies

\textbf{\ref{sec:sup_mmlm}} Details of Multi-modal Large Models

\textbf{\ref{sec:sup_qr}} More Qualitative Results

\textbf{\ref{sec:sup_fc}} Failure Cases

\textbf{\ref{sec:sup_fw}} Future Works

\section{Evaluation Metrics}
\label{sec:em}
In the era of competitive benchmarks, much research is evaluated on its ability to improve the scores. If benchmarks are using metrics to evaluate these scores which are skewed towards only certain aspects of a task, this will also steer research and models to focus more on these aspects. 
\subsection{mAP: mean Average Precision}
mAP (mean Average Precision) is a standard evaluation metric in image instance segmentation. It is the area under the precision-recall curve across multiple intersection-over-union (IoU) thresholds. The mAP metric has been extended to video instance segmentation, as proposed in \cite{vis}, where IoU computation differs from image instance segmentation because each instance contains a sequence of masks: 
\begin{equation}
\text{IoU}(\mathbf{G},\mathbf{P}) = \frac{\sum_{t=1}^T\left|\mathbf{m^G_t} \cap \mathbf{m^P_t}\right|}{\sum_{t=1}^T\left|\mathbf{m^G_t} \cup \mathbf{m^P_t}\right|}
\end{equation}
The proposed IoU computes the spatio-temporal consistency of ground-truth and predicted segmentation results. If the algorithm detects object masks but fails to track the objects across frames, the IoU score will be reduced.

However, mAP is not perfectly suited to our AVIS task, because it can be increased by producing many different predictions with low confidence scores and does not decrease even if non-sounding objects are predicted. Moreover, the threshold required for an instance to be considered a positive match is set high, resulting in lots of improvements in detection, association, and localization being overlooked by the evaluation metric. In addition, mAP mixes association, detection and localisation in a manner that does not allow for differentiation among error types.

\subsection{HOTA: Higher Order Tracking Accuracy}
HOTA (Higher Order Tracking Accuracy) \cite{hota} performs a bijective (one-to-one) matching at a detection level while scoring association globally over trajectories, which is designed for multi-object tracking task. This makes HOTA a balanced metric for measuring both detection and association. When applied to the AVIS task, it can penalize those models that predict non-sounding objects. 

A true positive (TP) refers to a matched pair of a ground-truth track set (gtDet) and a predicted detection set (prDet), for which the localisation similarity is greater than or equal to the threshold $\alpha$. A false negative (FN) is a gtDet that is not matched to any prDet. A false positive (FP) is a prDet that is not matched to any gtDet. The matching between gtDets and prDets is bijective within each frame. For a given TP, denoted as c, the set of TPAs is the set of True Positive Associates (TPs) which have both the same ground-truth id set (gtID) and the same predicted id set (prID) as c. For a given TP, c, the set of False Negative Associates (FNAs) refers to the set of gtDets with the same gtID as c, but that were either assigned a different prID as c, or no prID if they were missed. For a given TP, c, the set of False Positive Associates (FPAs) denotes the set of prDets with the same prID as c, but that were either assigned a different gtID as c, or no gtID if they did not actually correspond to an object. Having defined the concepts for measuring successes and errors in detection (TPs, FPs, FNs) and association (TPAs, FPAs, FNAs), the HOTA score can be defined as:
\begin{equation}
\begin{aligned}
& \text{HOTA}=\sqrt{\frac{\sum_{c \in\{\text{TP}\}} \mathcal{A}(c)}{|\text{TP}|+|\text{FN}|+|\text{FP}|}} \\
& \mathcal{A}(c)=\frac{|\text{TPA}(c)|}{|\text{TPA}(c)|+|\text{FNA}(c)|+|\text{FPA}(c)|}
\end{aligned}
\end{equation}

The HOTA can decompose into a separate detection accuracy score (DetA) and an association accuracy score (AssA) as follows:
\begin{equation}
\begin{aligned}
& \text{HOTA} =\sqrt{\text{DetA} \cdot \text{AssA}} \\
& \text{DetA} =\frac{|\text{TP}|}{|\text{TP}|+|\text{FN}|+|\text{FP}|} \\
& \text{AssA} =\frac{1}{|\text{TP}|} \sum_{c \in\{\text{TP}\}} \mathcal{A}(c)
\end{aligned}
\end{equation}

\subsection{FSLA: Frame-level Sound Location Accuracy}

\begin{algorithm}
\caption{The FSLA Evaluation Metric}
\label{algo}
\SetKwInOut{Input}{Input}
\SetKwInOut{Output}{Output}
\underline{function FSLA} $(\mathbf{M^P}, \mathbf{M^G}, \mathbf{C^P}, \mathbf{C^G}, \mathbf{ID^P}, \mathbf{ID^G})$\;
\Input{\small{A predicted mask set $\mathbf{M^P} = \{\mathbf{m^P_{i,l}}\}_{i=1,l=1}^{x,L}$.} \\
       \small{A labeled mask set $\mathbf{M^G} = \{\mathbf{m^G_{j,l}}\}_{j=1,l=1}^{y,L}$.} \\
       \small{A predicted class set $\mathbf{C^P} = \{\mathbf{c^P_i}\}_{i=1}^{x}$.} \\
       \small{A labeled class set $\mathbf{C^G} = \{\mathbf{c^G_j}\}_{j=1}^{y}$.} \\
       \small{A predicted id set $\mathbf{ID^P} = \{\mathbf{id^P_i}\}_{i=1}^{x}$.} \\
       \small{A labeled id set $\mathbf{ID^G} = \{\mathbf{id^G_j}\}_{j=1}^{y}$.} \\
       \small{The video frames $\mathbf{F}$. The video length $L$.} \\
       \small{$N_{fna}$, $N_{fsa}$, and $N_{fma}$ are the number of silent, single- and multi-sound-source frames.}}
\Output{\small{FSLA, FSLAn, FSLAs, FSLAm}}

\small{$N_{fnt}, N_{fst}, N_{fmt} \gets 0$} \\
\small{$S(x,y) = \text{{HungarianMatch}}(\mathbf{M^P}, \mathbf{M^G}, \mathbf{ID^P}, \mathbf{ID^G})$} \\
\For{\small{$\alpha \gets 0.05$ to $0.95$ step $0.05$}}{
    \For{\small{$l \gets 1$ to $L$ step $1$}}{
        \If{\small{$F_l$ is a silent frame}}{
            \small{$N_{fnt} \gets N_{fnt} + 1$}
        }
        \If{\small{$F_l$ is a single-sound-source frame}}{
            \If{$\mathbf{c^P_{i,l}} == \mathbf{c^G_{j,l}}$}{
                \If{\small{$S(i, j)$ and $IoU(\mathbf{m^P_{i,l}}, \mathbf{m^G_{j,l}}) > \alpha$}}{
                    \small{$N_{fst} \gets N_{fst} + 1$}
                }}
        }
        \If{\small{$F_m$ is a multi-sound-source frame}}{
            \If{\small{$\mathbf{c^P_{i,l}} == \mathbf{c^G_{j,l}}$}}{
                \If{\small{$S(i, j)$ and $IoU(\mathbf{m^P_{i,l}}, \mathbf{m^G_{j,l}}) > \alpha$}}{
                    \small{$N_{fmt} \gets N_{fmt} + 1$}
                }}
        }
    }
    \small{FSLAn($\alpha$) $\gets$ $N_{fnt} / N_{fna}$, FSLAs($\alpha$) $\gets$ $N_{fst} / N_{fsa}$} \\
    \small{FSLAm($\alpha$) $\gets$ $N_{fmt} / N_{fma}$} \\
    \small{FSLA($\alpha$) $\gets$ $(N_{fnt} + N_{fst} + N_{fmt}) / L$} \\
}
\small{FSLA $\gets$ $\overline{\text{FSLA}(\alpha)}$} \\
\small{FSLAn, FSLAs, FSLAm $\gets$ $\overline{\text{FSLAn}(\alpha)}$, $\overline{\text{FSLAs}(\alpha)}$, $\overline{\text{FSLAm}(\alpha)}$}
\end{algorithm}

Besides considering the above object-based metrics, we propose a novel measure, namely frame-level sound localization accuracy (FSLA), tailored to measure the proportion of frames that are correctly predicted by the model out of the total number of frames. Specifically, we first use the Hungarian algorithm \cite{kuhn1955hungarian} to determine a one-to-one matching between ground-truth and predicted detections. For each frame, it can be treated as correct frame if it satisfies the following conditions: 1) The number of sounding objects is correct; 2) The category of the sounding objects is correct; 3) The IoU (Intersection over Union) between the ground truth and the predicted sounding objects is greater than threshold $\alpha$. The final score is computed by averaging over all classes before averaging different $\alpha$ thresholds (0.05 to 0.95 in 0.05 intervals). The pseudo code of the FSLA metric is shown in Algorithm \ref{algo}. Compared to other metrics, our FSLA allows for easier localization of incorrect frames and offers a more intuitive explanation of the model's performance across different time periods. Additionally, it can be decomposed into a set of sub-metrics (FSLAn, FSLAs and FSLAm) which can be used for model evaluation in scenarios with no sound source, a single sound source, and multiple sound sources. This results in FSLA being able to guide how models can be improved, or understand where they are likely to fail when used.

\section{Implementation Details}
\label{sec:sup_id}
The audio and video frames are sampled at rates of 16 kHz and 1 FPS, respectively. For the image encoder, we attempt two different backbones, ResNet-50/101 \cite{resnet} and Swin-L \cite{swin}. For the audio encoder, we adopt VGGish \cite{vggish} pre-trained on AudioSet, with its parameters frozen during the training phase. Unless specified, the window size $W$ is set to 6, and both the number of frame queries and video queries are set to 100. Our model is implemented on top of the detectron2\footnote[1]{\footnotesize\url{https://github.com/facebookresearch/detectron2}} and trained on the proposed AVISeg dataset for 48,000 iterations with a batch size of 1. We use the AdamW optimizer and the step learning rate schedule. The initial learning rate is set to 1e-4 and reduced by a factor of 0.1 at 32,000 iterations. By default, the shorter side of frames are resized to 360 and 448 pixels during inference. The mask predictions are obtained without any post-processing, such as NMS. We keep predictions with a confidence threshold greater than 0.3. The experiments are conducted on 2 NVIDIA Quadro 6000 GPUs.

\section{More Ablation Studies}
\textbf{Impact of similarity loss and hyper-parameter setup.} As shown in Table \ref{tab: loss}, removing similarity loss yields a significant decrease across all metrics. This is because the model struggles to learn correct associations between object tokens and video queries, leading to feature misalignment, identity switches and tracking failures, especially for different instances of the same category. Additionally, we test several hyper-parameters and set $\lambda_{\text{sim}}=0.5$ as default, which achieves the best performance.

\label{sec:mas}
\begin{table}[h]
\begin{center}
\caption{Impact of similarity loss and hyper-parameter setup.}
\label{tab: loss}
\resizebox{\linewidth}{!}{
\begin{tabular}{c|ccc|ccc}
\shline
similarity loss   & $\lambda_{\text{sim}}=0.1$  & $\lambda_{\text{sim}}=0.5$ & $\lambda_{\text{sim}}=1.0$              & FSLA   & HOTA   & mAP  \\ \shline
\usym{2718}       &  &   &              & 32.71  & 52.45  & 35.77 \\ \shline
       & \usym{2714} &   &              & 38.97  & 59.92  & 38.22  \\
\usym{2714}       &  & \usym{2714}  &   & 42.78  & 61.73  & 40.57  \\
       &  &   & \usym{2714}             & 42.08  & 61.63  & 40.49  \\
\shline
\end{tabular}}
\end{center}
\end{table}
\vspace{-1.3em}

\section{Details of Multi-modal Large Models}
\label{sec:sup_mmlm}

\subsection{Sam4AVS} \label{sec:sam4avs}
\begin{figure*}[!htp]
\centering
\includegraphics[width=1.0\textwidth]{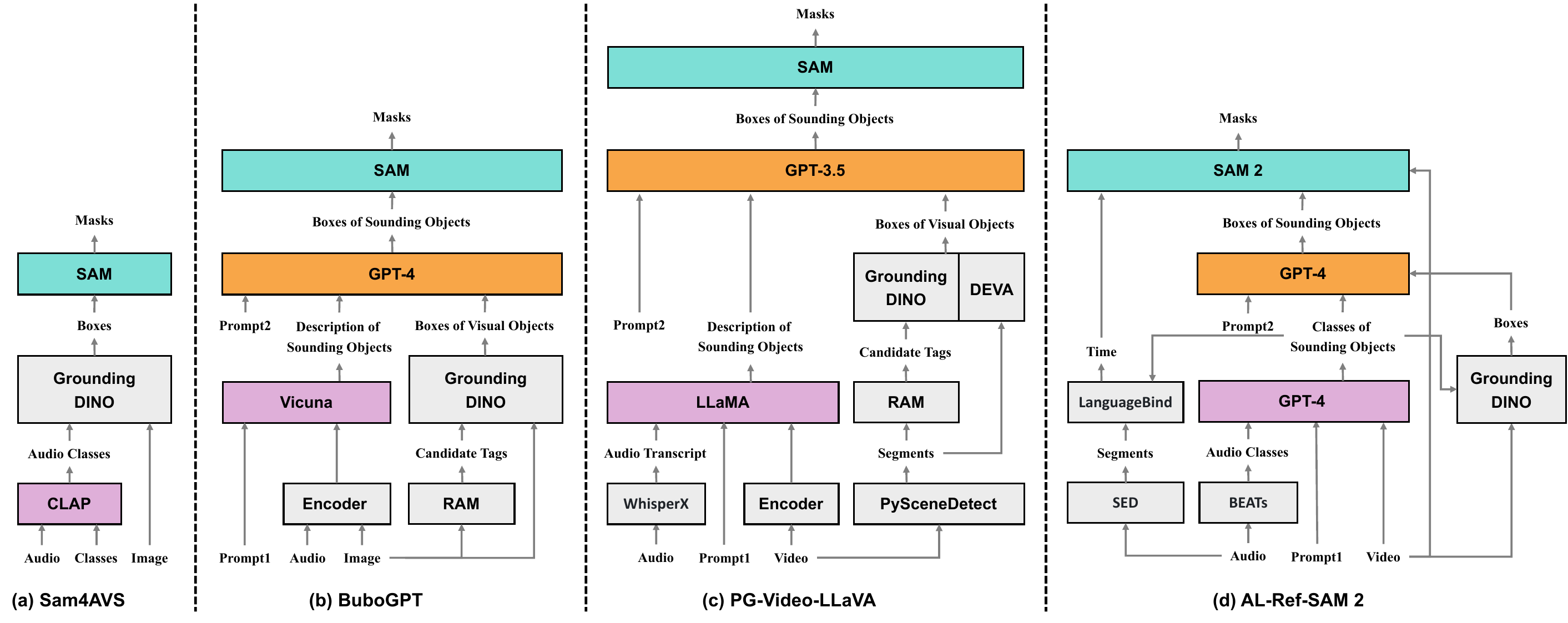}
\put(-458,5.5){\scriptsize ~(\S\ref{sec:sam4avs})}
\put(-364,5.5){\scriptsize ~(\S\ref{sec:bubogpt})}
\put(-220,5.5){\scriptsize ~(\S\ref{sec:pg})}
\put(-65,5.5){\scriptsize ~(\S\ref{sec:alref})}
\caption{Pipeline comparison of multi-modal large models, including (a) Sam4AVS \cite{sam4avs}, (b) BuboGPT \cite{bubogpt}, (c) PG-Video-LLaVA \cite{pg-video-llava}, and (d) AL-Ref-SAM 2 \cite{al-ref-sam}. \textcolor{mmfusion}{\textbf{Multi-modal fusion module}} aligns audio-X modalities and outputs classes or descriptions of sounding objects. \textcolor{assist}{\textbf{Assistant module}} leverages the reasoning capabilities of large language models to retrieve matched sounding objects. \textcolor{seg}{\textbf{Segmentation module}} adopts an off-the-shelf visual grounding pipeline to localize sounding objects and generate corresponding fine-grained masks.}
\label{fig_sup_model_comp}
\end{figure*}

\textbf{Model.} As shown in Figure \ref{fig_sup_model_comp} (a), Sam4AVS \cite{sam4avs} leverages the large-scale audio-language model CLAP \cite{clap} to classify the input audio. For a single-source audio, the class name with the highest score is selected, while for a multi-source audio, the two highest-scoring class names are chosen. The predicted class names are input into Grounding DINO \cite{groundingdino} to generate box predictions, and these boxes are then utilized to query SAM \cite{sam} for mask generation.

\textbf{Experiment.} We reproduce Sam4AVS and make it suitable for the AVIS task. Specifically, we divide each audio into multiple 1-second segments and feed them into CLAP separately. Then, we select the class name with top-1 score to generate masks for each video frame. Furthermore, masks of the same category throughout the entire video are considered to belong to the same object.

\textbf{Problem.} Only using audio information to predict the category of sounding objects proves insufficient and unreliable in complex scenarios. For instance, humans can imitate the sound of a cat meowing, and both cars and airplanes may generate similar engine sounds. Sam4AVS neglects visual cues, potentially leading to inaccurate classification of sounding objects. When provided with a class name, Sam4AVS tends to segment all objects belonging to the predicted class, rather than those sounding ones. Additionally, Sam4AVS processes images individually, which prevents it from establishing temporal correlations or tracking instances of sounding objects.

\subsection{BuboGPT} \label{sec:bubogpt}
\textbf{Model.} As shown in Figure \ref{fig_sup_model_comp} (b), BuboGPT \cite{bubogpt} aligns audio-vision-language modalities while leveraging a large language model to generate description of sounding objects. It employs an existing visual grounding pipeline to find the above sounding objects described above in an image and output their final masks. More specifically, BuboGPT uses ImageBind \cite{imagebind} as the audio encoder, BLIP-2 \cite{blip2} as the vision encoder and Vicuna \cite{vicuna} as the large language model. BuboGPT first aligns audio or visual features with language by training the modality Q-Former \cite{blip2} and linear projection layer on audio or image caption datasets, respectively. Subsequently, it conducts multi-modal instruct tuning on a large instruction-following dataset, prompting Vicuna to generate description of sound source. The prompt template, i.e., prompt1 depicted in Figure \ref{fig_sup_model_comp} (b), is defined as follows:
\begin{lstlisting}
<Vision><ModalityHere></Vision> <Audio><ModalityHere></Audio> Please find the source that emits the given sound in this image.
\end{lstlisting}
\noindent To explore the relationships between different visual objects and descriptions of sound source, BuboGPT adopts an off-the-shelf visual grounding pipeline based on SAM \cite{sam}. This pipeline consists of four modules: 1) a tagging module RAM \cite{ram} to produce multiple text tags/labels that are relevant to the input image; 2) a grounding module Grounding DINO \cite{groundingdino} responsible for localizing a bounding box in the image corresponding to each tag/label; 3) an entity-matching module GPT-4 \cite{gpt4} that leverages the reasoning capabilities of the large language model to retrieve matched entities from tags and image descriptions; 4) a segmentation module SAM \cite{sam} designed to get fine-grained masks. The prompt template of the entity-matching module, i.e., prompt2 depicted in Figure \ref{fig_sup_model_comp} (b), is defined as follows:
\begin{lstlisting}
You are a helpful assistant. Now I will give you a list of entities and give you a paragraph or sentence. You need to first extract the entity given in the text and then find the corresponding entity having similar or identical meanings in the given list. Find all the pairs. Are you clear? let us think step by step. The extracted entities must come from the given text and the corresponding entity must come from the given list. If multiple entities can be linked to the same span of text or vice versa, just keep one and do not merge them. Here is an example: <List>['dog','sheepdog','grass','chase sheepdog','field','field park','grassy','corgi','brown dog','brown','park']</List> <Text>A brown dog running in the grassy field</Text> The answer is: brown dog - brown dog \n grassy field - field
\end{lstlisting}

\textbf{Experiment.} We reproduce BuboGPT and make it suitable for the AVIS task. Specifically, we split each video into multiple non-overlapping visual and audio snippet pairs, where each snippet spans 1 second. BuboGPT takes an image and the corresponding 1-second audio as input, and generate masks for each video frame. Furthermore, masks of the same tag/category throughout the entire video are considered to belong to the same object.

\textbf{Problem.} Compared to Sam4AVS, BuboGPT integrates both audio and visual information to classify and localize sounding object instances, resulting in more accurate sound source localization. However, it still only process one image at a time, which prevents it from establishing temporal correlations or tracking instances of sounding objects. Moreover, RAM predicts tags/categories rather than providing detailed descriptions of the objects. Therefore, the entity-matching module struggles to differentiate between different object instances of the same category.

\subsection{PG-Video-LLaVA} \label{sec:pg}
\textbf{Model.} As shown in Figure \ref{fig_sup_model_comp} (c), PG-Video-LLaVA \cite{pg-video-llava} transcribes audio cues into texts and extracts spatio-temporal features from videos. Then they are input into a large language model to generate description of sounding objects. Finally, PG-Video-LLaVA uses an off-the-shelf tracker along with a visual grounding module, allowing it to spatially segment sounding objects in videos according to the generated descriptions. Specifically, PG-Video-LLaVA takes video frames as input and employs the CLIP \cite{clip} visual encoder to extract video features by averaging frame-level features across temporal and spatial dimensions. For the audio modality, PG-Video-LLaVA utilizes WhisperX \cite{whisperx}, a speech recognition system, to detect voice activity and generate audio transcripts. The integration of the audio transcript with the video features is executed in the large language model LLaMA \cite{llama} through a carefully designed prompt template, i.e., prompt1 depicted in Figure \ref{fig_sup_model_comp} (c):
\begin{lstlisting}
You are PG-Video-LLaVA, a large vision-language assistant. You are able to understand the video content that the user provides, and assist the user with a variety of tasks using natural language. Your task is to find the source that emits the given sound in this video. <Video-Tokens> The noisy audio transcript of this video is: <Audio-Transcript>
\end{lstlisting}
\noindent After obtaining descriptions of sounding objects from LLaMA, these are employed for grounding within the corresponding video frames. Key noun phrases are extracted from the generated text via GPT-3.5, focusing on the category of sounding objects. The prompt template of GPT-3.5, i.e., prompt2 depicted in Figure \ref{fig_sup_model_comp} (c), is similar to BuboGPT. Simultaneously, an image tagging model, RAM \cite{ram}, tags visual elements in each frame, constructing a detailed map of the video content. The video is segmented into smaller parts using PySceneDetect, based on changes in scene composition. In each segment, a grounding ensemble, composed of GroundingDINO \cite{groundingdino}, DEVA \cite{deva}, and SAM \cite{sam}, employs the image tags to generate segmentation masks and tracking IDs for the identified visual elements. The visual cues from these segmentation masks are subsequently matched with the textual noun phrases through CLIP \cite{clip}. This matching process links the text to the corresponding visual elements in the video.

\textbf{Experiment.} We reproduce PG-Video-LLaVA and make it suitable for the AVIS task. Specifically, each noun phrase from GPT-3.5 serves as an instance and is then input into the grounding module to generate segmentation masks throughout the entire video.

\textbf{Problem.} PG-Video-LLaVA extends image-based large multi-modal models to the video domain, and provides a more accurate understanding of video content compared to Sam4AVS and BuboGP. Nevertheless, it can only describe what the sounding object in the video is but cannot pinpoint the exact time intervals for each sounding object. Moreover, for each video, its feature are obtained by simply averaging image features, which may result in the loss of some valuable information. For each audio, PG-Video-LLaVA only identifies speech segments, filtering out non-speech audio components (e.g., music, machine or animal sounds), and transcribes the speech into text. In addition, RAM predicts tags/categories rather than providing detailed descriptions of the objects. Therefore, GPT-3.5 struggles to differentiate between different object instances of the same category.

\subsection{AL-Ref-SAM 2} \label{sec:alref}
\textbf{Model.} As shown in Figure \ref{fig_sup_model_comp} (d), AL-Ref-SAM 2 \cite{al-ref-sam} employs an intuitive three-stage pipeline for the audio-visual segmentation task: 1) extract reference information from the multi-modal input, 2) identify the sounding object in the initial frame based on the extracted reference, and 3) segment the identified sounding object throughout the entire video. Specifically, AL-Ref-SAM 2 applies an audio classifier, BEATs \cite{beats}, to categorize the audio clip. 
\begin{lstlisting}
The image is composed of multiple frames from a video spliced from left to right, and the frame number is marked with a circle in the upper left corner of each frame. Using an audio classification model, we obtained the audio labels with the highest confidence in the video: {$OBJ_1$,$OBJ_2$,...,$OBJ_k$}. Please process these audio labels based on the content of the image, filtering out audio labels that do not exist in the video or are abstract labels that cannot be associated with specific objects. Additionally, merge audio labels that represent the same object. Then, according to the retained audio labels, output the category of one or more objects in the video that may be making sounds in a list surrounded by [].
\end{lstlisting}

\begin{lstlisting}
I have input an image stitched together from frames of a video, each frame is marked with an ID in the upper left corner. Please first describe in detail the events happening in the video and then help me select the single frame that best demonstrates the \"{reference}\" and may result in a good segmentation result of the object previously described, and return their IDs in the upper left corner to me in a list surrounded by [].

The above content is an image that contains sampled frames of a video, with the frame numbers labeled in the top-left corner. In the {$p_f$} frame, three objects are marked with colored boxes: {$bbox_1$,$bbox_2$,$bbox_3$}. Please follow these steps:
1. Describe the Scene: Describe the video and each frame. Describe each object in the frame.
2. Describe the Objects within Each Box: Describe the objects in the above boxes and their relationships.
3. Analyze the Provided Description: Given the description \"{reference}\" and analyze its syntax, identifying the main object described in the sentence. Adhere to syntax analysis principles, and do not assume that an object is the main subject simply because it an has extensive description. This analysis will help you distinguish the box that needs to be selected from the image.
4. Identify the Object that Best Matches the Description:
Ensure you select the precise bounding box of the referring object by following these tips: Include only the main object described, excluding other objects. Include the whole main object. Do not include other objects mentioned in the description that are not the main object.
5. Output the Result: Output the single number in list [] format.
\end{lstlisting}

\noindent To avoid the disturbance presence of background noise and the ambiguity of audio information, it integrates visual context and leverages the vision-language understanding capabilities of existing large multi-modal model, GPT-4 \cite{gpt4}, to accurately identify the categories of the actual sounding objects present in the video. The prompt template, i.e., prompt1 depicted in Figure \ref{fig_sup_model_comp} (d), is defined as mentioned above. Since the selected referent may be silent in certain frames, AL-Ref-SAM 2 further utilizes sound event detection (SED) to segment the whole audio clip and filter out silent frames from the generated mask sequence. Then, GPT-4 processes the identified categories and video clip to identify a high-quality box of the referent in a specific frame where the referent clearly appears. The prompt template, i.e., prompt2 depicted in Figure \ref{fig_sup_model_comp} (d), is defined as above. Finally, the selected bounding box serves as the pivot box to prompt SAM 2 \cite{sam2} to segment the referent and propagate its mask forward and backward through the entire video. 

\textbf{Experiment.} We reproduce AL-Ref-SAM 2 and make it suitable for the AVIS task. Specifically, each category is considered as a individual object instance.

\textbf{Problem.} Compared to PG-Video-LLaVA, AL-Ref-SAM 2 is capable of determining the exact time intervals during which objects emit sound. However, it cannot distinguish between different object instances of the same category, because BEATs and GPT-4 only output the category of the audio rather than a description of sounding objects.

\section{More Qualitative Results}
\label{sec:sup_qr}
\begin{figure*}[!htp]
\centering
\includegraphics[width=1.0\textwidth]{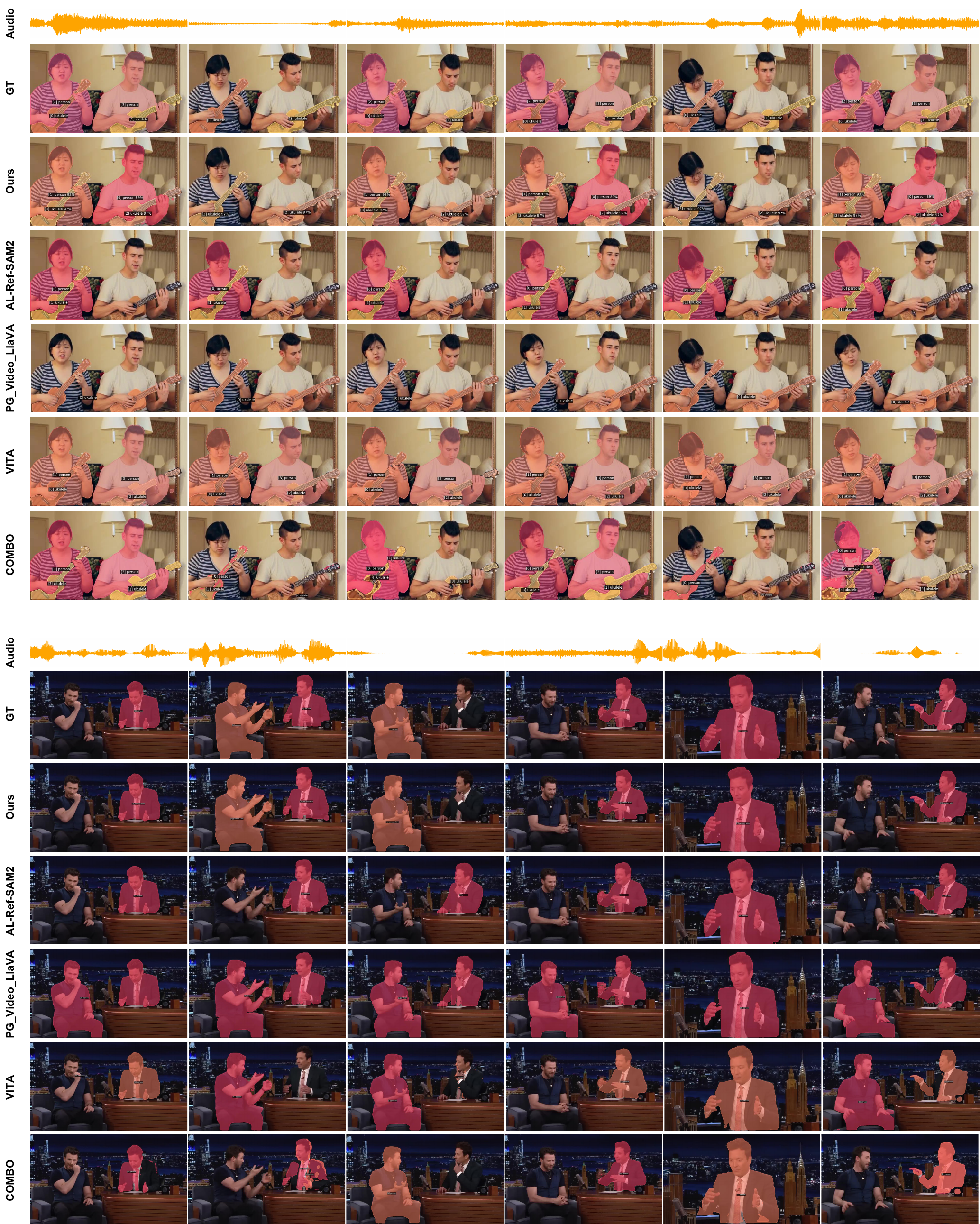}
\caption{Qualitative comparison of our model with VIS (VITA), AVSS (COMBO) and multi-modal large models (PG-Video-LLaVA and AL-Ref-SAM 2) on Music (Top) and Speaking (Bottom) scenarios.}
\label{fig_sup_results1}
\end{figure*}

\begin{figure*}[!htp]
\centering
\includegraphics[width=1.0\textwidth]{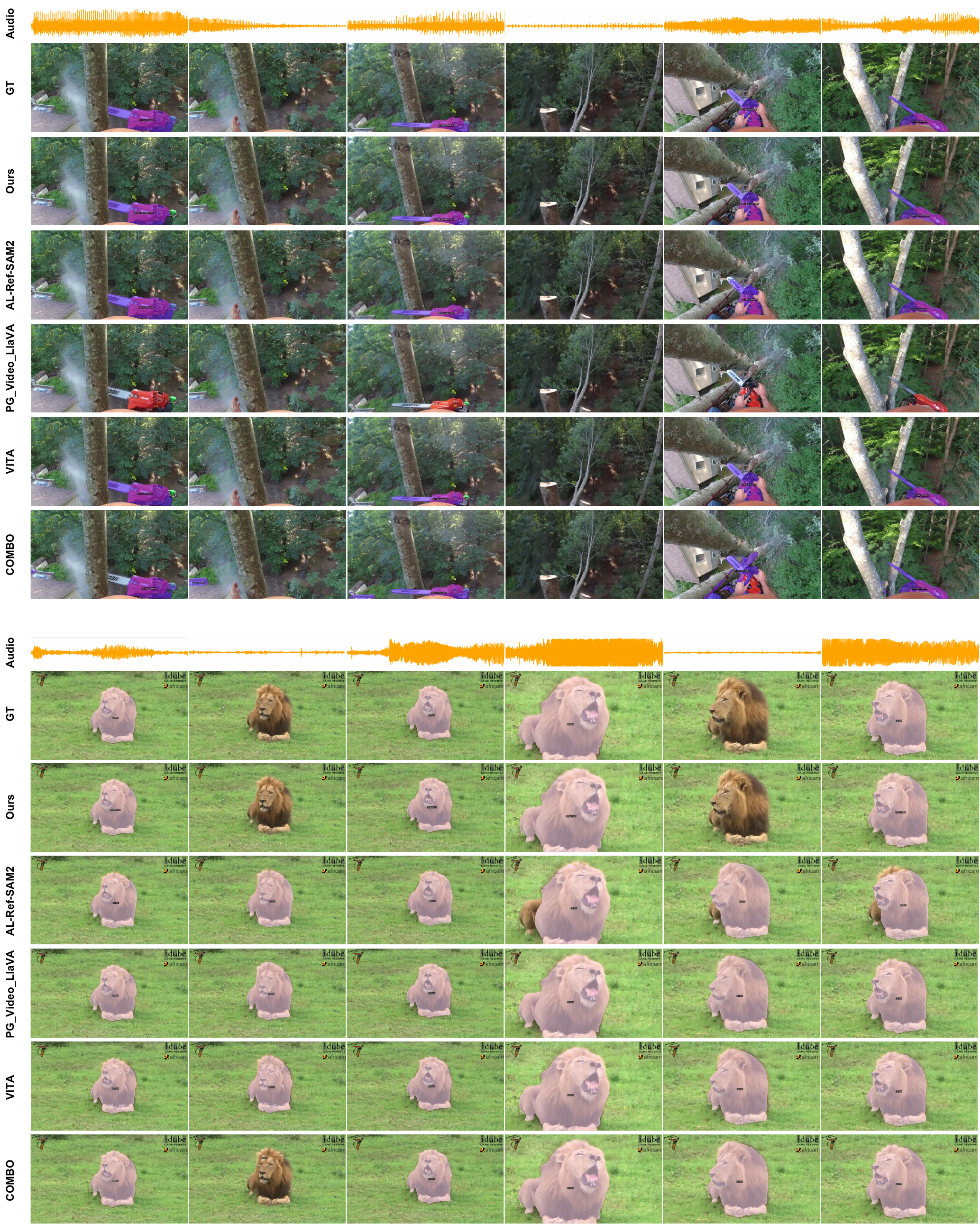}
\caption{Qualitative comparison of our model with VIS (VITA), AVSS (COMBO) and multi-modal large models (PG-Video-LLaVA and AL-Ref-SAM 2) on Machine (Top) and Animal (Bottom) scenarios.}
\label{fig_sup_results2}
\end{figure*}

As shown in Figure \ref{fig_sup_results1} and Figure \ref{fig_sup_results2}, we provide some qualitative comparisons with other methods on 4 scenarios. \textbf{1)} Video instance segmentation methods (e.g., VITA \cite{vita}) can accurately segment and track objects, but fails to determine when these objects are producing sound, e.g., ``person'' in Figure \ref{fig_sup_results1} and ``lion'' in Figure \ref{fig_sup_results2}, due to the absence of audio input. \textbf{2)} With the help of audio information, audio-visual semantic segmentation methods (e.g., COMBO \cite{combo}) are capable of correctly localizing the sound source in most cases. However, such methods show difficulties in processing long sequences, which may result in multiple identity switches in tracking, e.g., ``person'' in Figure \ref{fig_sup_results1}. \textbf{3)} Multi-modal large models (e.g., PG-Video-LLaVA \cite{pg-video-llava} and AL-Ref-SAM 2 \cite{al-ref-sam}) serve audio as a form of language and leverage foundation models to achieve audio-referred visual grounding. As discussed in Section \ref{sec:sup_mmlm}, these methods not only fail to distinguish between different object instances of the same category, e.g., ``person'' in Figure \ref{fig_sup_results1}, but also struggle to determine the exact time intervals for each sounding object, e.g., ``lion'' in Figure \ref{fig_sup_results2}.

\begin{figure*}[!htp]
\centering
\includegraphics[width=1.0\textwidth]{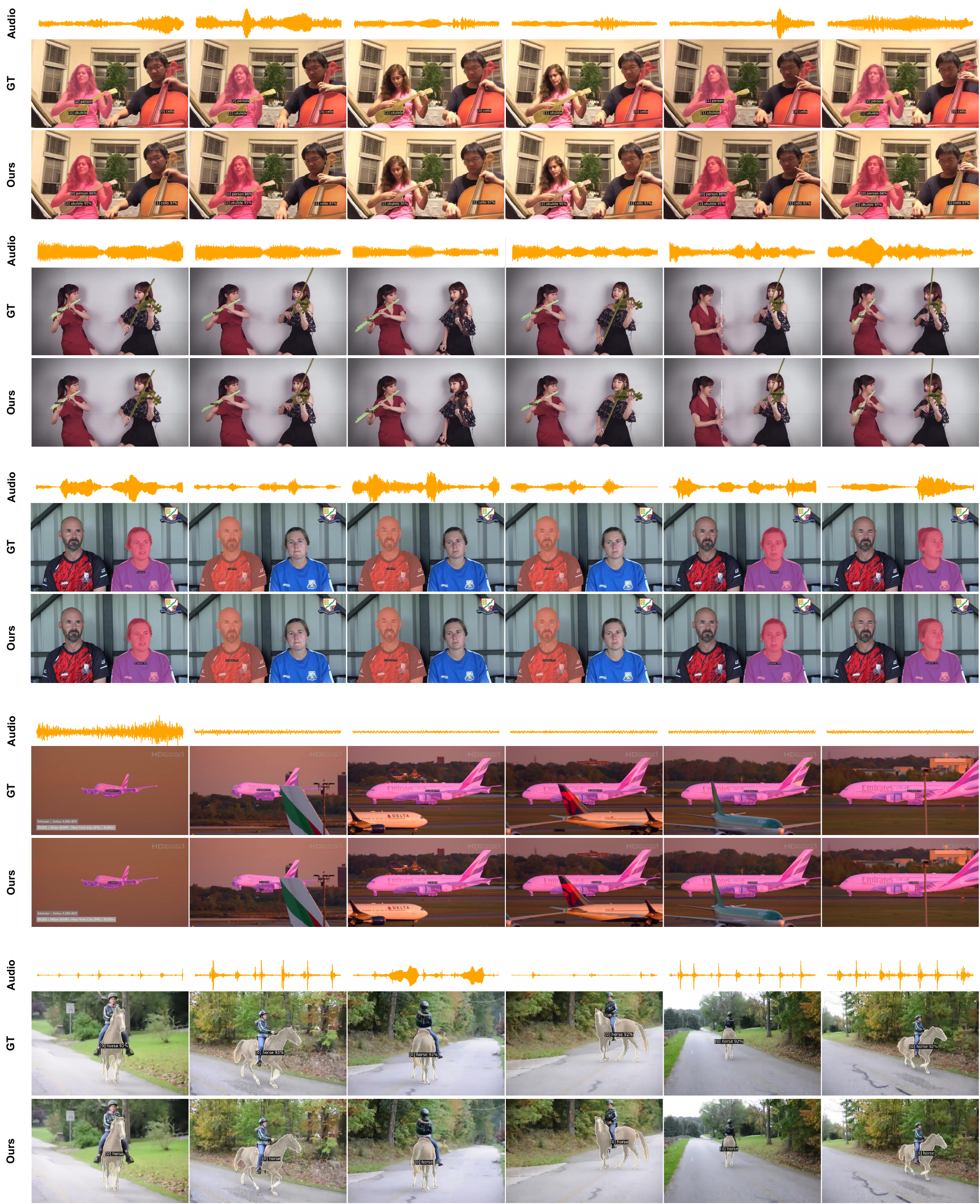}
\caption{More visual results of our baseline model on AVISeg dataset from four scenarios. Each row have six sampled frames from a video sequence. Zoom in to see details.}
\label{fig_sup_results3}
\end{figure*}

In addition, we show more visual results of our baseline model in Figure \ref{fig_sup_results3}. Our model accurately localizes sound sources, segments sounding objects, and determines when they are emitting sound.

\section{Failure Cases}
\label{sec:sup_fc}
\begin{figure*}[!htp]
\centering
\includegraphics[width=1.0\textwidth]{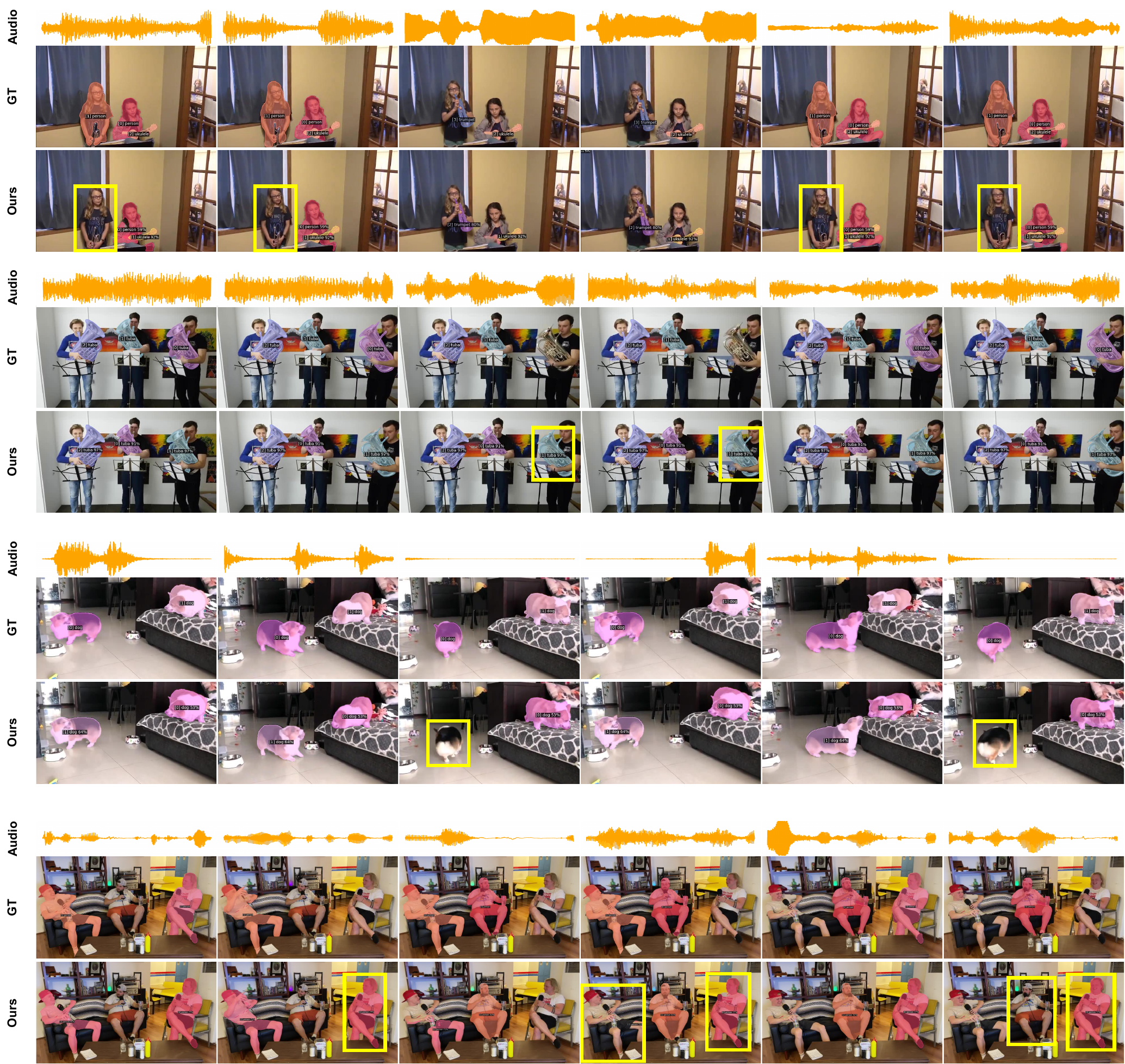}
\caption{Failure cases of our baseline model on AVISeg dataset. Each row has six sampled frames from a video sequence. The yellow boxes indicate the incorrect segmentation regions. Zoom in to see more details.}
\label{fig_sup_results4}
\end{figure*}

Figure \ref{fig_sup_results4} displays additional failure cases of our model on the AVISeg dataset. We observe that inaccurate sound source localization tends to occur in complex multi-source scenarios, especially when multiple objects within the same category emit sound, e.g., two ``girls'', three ``tubas'', two ``dogs'' and three ``men'' in Figure \ref{fig_sup_results4}. This because audio signals from homogeneous sounding objects often exhibit similarity and indistinguishable, making them complicating the alignment with visual content. It motivates us to explore how to more effectively disentangle high-density audio signals and establish robust correspondences between audio and visual contents in complex multi-source scenarios and long video sequences.

\section{Future Works}
\label{sec:sup_fw}
As a pioneering work, the current approach is not perfect and thus leaves much room for improvement, which we summarize below:

\textbf{1) Long-range temporal modeling.} % sink attention
% The current method faces challenges when applied to infinite length audio-visual input.
Recent work by StreamingLLM \cite{streamingllm} introduces the concept of ``attention sinks'', additional initial tokens that consistently participate in attention computations during sliding window processing. This enables models trained with finite attention windows and generalize to infinite-length sequences without requiring further fine-tuning. Adopting this technique could potentially enhance long-range consistency and improve performance across extended audio-visual sequences.

\textbf{2) Audio decoupling and audio-visual fusion.} 
As discussed in Section \ref{sec:sup_fc}, our model’s performance may be limited in scenarios where multiple objects of the same category are producing sound. To better associate mixed-source audio with visual objects, product quantization \cite{productquantization, combo} can be considered to decompose the mixed audio semantics into several disentangled single-source semantics with noise suppression. This approach has the potential to provide a more compact and robust audio representation for audio-visual interaction, especially in complex scenarios.

\textbf{3) Online audio-visual segmentation.} 
Many recently introduced methods have demonstrated promising performance for audio-visual segmentation tasks. However, they are restricted in real-time applications as they operate offline, requiring the entire video to be processed before the predictions. Therefore, developing online methods that processes video frames sequentially, without access to future frames, will be an important topic.

\textbf{4) Prompt engineering and instruction tuning.} 
With the help of large language models, existing multi-modal large models (MMLMs) exhibits impressive audio-visual understanding abilities. Nevertheless, they are far from satisfactory in fine-grained audio-referred visual grounding tasks, especially in instance-aware sound source localization and long videos. By carefully designing the text prompts or fine-tuning on the AVISeg-based instruction-tuning dataset, MMLMs can produce more accurate responses and detailed descriptions of sounding objects.

\end{document}